\documentclass[11pt]{article}
\usepackage[margin=1in]{geometry}
\usepackage{hyperref}
\usepackage{url}
\usepackage{amsmath}
\usepackage{booktabs}
\usepackage{graphicx}

\usepackage[numbers,sort&compress]{natbib}

\usepackage[nameinlink,noabbrev]{cleveref}

\newcommand{\atemp}{\tau}

\title{Concentration, Not Uncertainty: Why Targeted Synthetic Data Doesn't Help Camouflaged Object Detection}

\author{
    Akshat Dobhal \qquad
    Sanjay Singh \qquad
        \\
    Manipal Institute of Technology, Manipal Academy of Higher Education, Manipal, India \\
    \\
    \texttt{akshat.dobhal@learner.manipal.edu, sanjay.singh@manipal.edu}
}
\date{October 2026}

\begin{document}

\maketitle

\begin{abstract}
Camouflaged object detection requires pixel-accurate masks, but obtaining such annotations is slow and costly, making synthetic training images an attractive alternative. Under a fixed generation budget, however, it remains unclear which real-image regions to target for synthetic data generation. We study an uncertainty-guided generation strategy that clusters the unlabelled real images, identifies clusters on which the model is least certain, allocates synthetic generation toward those clusters, and iteratively retrains the model. Across 103 training runs, uncertainty-based targeting does not outperform random allocation. Five independent controls further show that this null result is not an artifact: targeted training sets are measurably different from random sets, but the difference is explained by concentrating the generation budget rather than by where uncertainty is concentrated, as every concentration rule we test reproduces the effect and, on boundary accuracy, so does aiming at the clusters the model was most certain about. Separately, we find substantial data contamination in CHAMELEON, with 50 of its 76 images duplicated from training data despite the standard overlap check reporting zero overlap. Together, these results show that, under a fixed synthetic-data budget, budget concentration, not uncertainty-based targeting, accounts for the observed training-set effects.
\end{abstract}

\section{Introduction}
\label{sec:intro}
Camouflaged object detection is a task aimed at generating an accurate pixel-level outline of an animal that evolved to blend into its surroundings. Its training masks are unusually expensive for exactly that reason: an annotator must
decide where a boundary lies when the point of the object is to conceal its boundary. One standard response to the resulting scarcity is to manufacture more data. A generator takes a real
cutout object and paints a fresh background around it, producing a new labeled pair at no
annotation cost.

Generation is not free either, and every generated image has to be trained on. So if you generate
a thousand of them, which thousand? Our answer is to let the model guide its own data generation: After initial training, we evaluate the model on unlabelled real images, identify the regions where highest uncertainty is observed, and target our synthetic data generation toward those regions. The methods of that loop are
Standard and uses uncertainty sampling from active learning~\cite{settles2009active}, a student--teacher
disagreement score~\cite{tarvainen2017mean}, clustering over self-supervised
features~\cite{oquab2024dinov2} so the budget spreads rather than piling onto near-duplicates. The
composition is ours, and we failed to make it function.

A targeted training set really is measurably different from a random one of the same size: a classifier, shown images from each, learns to tell which set an image came from even images it never saw. But it is picking up concentration, not aim. Targeting piles the budget onto a few clusters, and any rule that piles it the same way works just as well: point the same lopsided budget at the wrong clusters and the classifier still wins. Only removing the unevenness itself defeats it. What looked like the effect of targeting was the effect of concentrating, the
comparison that settles it is not a random set, but one with the same lopsided shape
aimed at nothing in particular.

In terms of trained accuracy, modifying the direction of the uncertainty signal shows no measurable effect that this design can resolve. That null is
bounded rather than absolute, and a positive control confirms that the measurement detects a real effect.
Substituting a perfect score using a test-set error for ours changes nothing. Extending training to accommodate extra images without reducing steps on real data shows no improvement. Moreover, re-running the decisive comparison at eight
seeds per scenario leaves the gap \emph{smaller} than it was at three. Separately, we expose a major flaw in the standard CHAMELEON test set, where 50 of its 76 images are direct copies from COD10K and CAMO training sets. Standard duplicate checks fail to flag these because they were re-saved in different formats, eliminating byte-level matches. The main contributions of this paper are:

\begin{itemize}
\item \textit{A control} that separates aiming a generation budget from merely concentrating it. Because a steered selection differs from random in two ways at once where it points and how unevenly it is spread a random baseline conflates them. Ours fixes the shape and scrambles only the signal; our own method does not survive it.
\item \textit{A bounded null on the direction}, defended by five independent controls: a positive
control, a perfect-score oracle, an unpinned optimization schedule, a same-shape shuffle, and a seed
expansion rather than asserted once.
\item \textit{Three checks run before any training} and precede any cluster-wise acquisition
policy: whether the target data clusters at all, what the funded clusters are made of, and how much of the endpoint's error sits between clusters rather than within them.
\item \textit{CHAMELEON is $50$ of $76$ images ($65.8\%$) training data} for models trained on
COD10K, while exact hashing reports $0$ overlap; we release a re-encode-tolerant detector, the
confirmed pair list and a clean evaluation protocol.
\end{itemize}

\subsection{Related Work}
\label{sec:related}
This paper asks whether an \emph{acquisition policy}, a rule for
deciding which synthetic images to generate under a fixed budget can improve a pipeline that is
otherwise held fixed. The task it does that on is camouflaged object detection, which asks for a
pixel-accurate outline of an object whose color and texture are built to defeat exactly that;
\cite{fan2023advances} surveys its architectures, datasets, and evaluation conventions.

A prediction is scored by how well its mask matches the true one. Our primary measure is the \emph{S-measure} $S_\alpha$~\cite{fan2017structure}, which rewards getting an object's \emph{shape} right rather than counting correct pixels, and which runs
from $0$ to $1$ with higher being better. The scale matters for reading everything that follows. Every
camouflaged object detector in this paper scores near $0.70$, so the differences the paper turns on are small fractions
of the range these models actually occupy. Three words then recur. An \textit{arm} is one training
condition; all arms share a base pool of labeled images and differ only in which images are
appended to it. A \textit{seed} is one training run of one arm, repeat runs of the same arm do
not return the same score, and the size of that seed-to-seed spread is what any real difference has
to beat. An \textit{endpoint} is a held-out test set from which a verdict is read. Finally, the
\textit{reference effect} is the improvement this repository reproduced for the published method it
builds on, $0.7030 \rightarrow 0.7172$, or $+0.0142$. That is the size of effect this design had to
be able to resolve, and the yardstick every sensitivity claim in this paper is measured against. The closest are LAKE-RED~\cite{zhao2024lakered}, whose background generation and compositing we use unchanged, and
S2R-COD~\cite{luo2025s2rcod}, whose synthetic-to-real self-training protocol we follow. Relative to
both, the only thing we add is an allocation layer that chooses \emph{which} generated images to
append; the generator, the budget, and the training procedure are left alone. Synthetic-augmentation methods such as DatasetDM~\cite{wu2023datasetdm} address how to construct labeled data at all, which is a different question from how to spend a fixed amount of it.

Every part of it is standard. Uncertainty sampling comes from active learning~\cite{settles2009active}, except that our query is answered by a generator rather than by a human annotator; the uncertainty is student-teacher disagreement of the
kind mean-teacher training already minimises~\cite{tarvainen2017mean}; and $k$-means over frozen
self-supervised features~\cite{oquab2024dinov2,radford2021learning} keeps the budget from piling
onto near-duplicates, in the spirit of core-set selection~\cite{sener2018active}. Other
constructions exist: deep ensembles~\cite{lakshminarayanan2017simple},
MC-dropout~\cite{gal2016dropout}, BADGE~\cite{ash2020deep}, and Section~\ref{sec:why} measures
an ensemble and predictive entropy alongside ours, but we propose none. What is contributed here is a controlled test of the composition. 

Three recent systems share the premise that a generation budget should be steered rather than spent uniformly.
GAUDA~\cite{frisch2025gauda} conditions a latent diffusion model on the classes a Bayesian model is
least certain about; DisCL~\cite{liang2025discl} steers image-guidance strength along a curriculum;
SynQuE~\cite{chen2025synque} ranks whole synthetic datasets using only unannotated real data.
\textit{None of the three runs a control that holds the allocation's shape fixed while destroying
the signal inside it}, and that turns out to be the control that decides the question, because a
steered selection differs from a random one in two ways at once, where the budget points, and how
unevenly it is spread. Appendix~\ref{app:related} places these systems side by side and covers
benchmark definitions and generative alternatives. Section~\ref{sec:method} specifies the loop we
built and the arms that separate those two things.

\section{Methods}
\label{sec:method}
In this section, we introduce our approach of uncertainty guided generation in the loop. We train a model on a labeled set of images, evaluate on the unlabeled
real images, and find the ones it handles least confidently. Then we ask a generator for
new images resembling those, add them to the training set, and train again. What follows is our
instantiation of that idea, specified exactly, because a null result is only as informative as the
thing it tested.

We train on a \emph{synthetic data source pool} of $4447$ labelled image and mask pairs
and want the model to work on a \emph{target pool} of $4040$ unlabelled real images, with verdicts
read from two held-out endpoints: COD10K-test decides, and NC4K is reported. The target pool is not
one distribution but a mixture of $3040$ COD10K images and $1000$ from CAMO, which matters later
(Section~\ref{sec:why}). Training runs in two rounds: a student trains alongside a teacher that is a slowly updated running
average of its own weights, target images that the two already agree on are appended between rounds with
the teacher's prediction as their label, and round two retrains from scratch on the enlarged pool.
Two properties the phrase ``closed loop'' hides both bound what follows; the round count is a
fixed constant, not a convergence criterion, and the generator renders its pool once, offline, and
is never invoked during training.

A fixed generation budget has to be spent somewhere, and our rule spends it where the model is weakest. \textit{(i) Group the target images}, so that
``weak'' can name a region of the data rather than a single image: $k$-means into $k = 75$ clusters
over L2-normalized self-supervised embeddings, fitted once and never refitted. \textit{(ii) Score
each image for uncertainty.} We need a number, computable without labels, that says how unsure the model is; the natural candidate is how far the student and its own running average disagree, which
is available precisely because the consistency loss is already trying to reduce it:
\begin{equation}
\mathrm{ES}(x) \;=\; a \,\big\| \nabla \sigma(s(x)) - \nabla \sigma(t(x)) \big\|_{1}
\;+\; b \, \mathrm{BCE}\big( \sigma(s(x)),\, \sigma(t(x)) \big),
\label{eq:es}
\end{equation}
with student and teacher logits $s,t$, sigmoid $\sigma$, Sobel gradient magnitude $\nabla$, binary
cross-entropy BCE, and
configured weights $a, b$ in Appendix~\ref{app:method}. What this buys is a score needing no ground
truth, computable on exactly the images the model will be judged on; a cluster's score is the mean
over its members. Its dominant term is the distance between \emph{gradient} magnitudes, which makes it a boundary-focused signal; a fact Section~\ref{sec:concentration} returns to. \textit{(iii) Turn scores into a budget split.} The budget is spread by a softmax over the cluster scores in units of their across-cluster standard deviation, at temperature $\atemp = 1.0$ (Appendix~\ref{app:method}); $\atemp$ sets how sharply the
budget piles onto the worst clusters.
\textit{(iv) Fill each quota} by drawing the foregrounds whose masked cutouts sit closest to the
funded cluster's centroid.

\begin{table}[t]
\caption{Every arm shares one base pool and differs only in what is appended to it. The last two columns are the axes of Figure~\ref{fig:arms}: how concentrated the allocation is, and where it points. The C-family are permutations of a single allocation, so they hold concentration fixed and differ only in direction; the digits in C10 record the allocation temperature,
$\atemp = 1.0$. None of A0, A2 and MT has an allocation to place.}
\label{tab:arms}
\begin{center}
\small
\setlength{\tabcolsep}{4pt}
\begin{tabular}{@{}llll@{}}
\toprule
Arm & What it appends & Concentration & Direction \\
\midrule
A0 & nothing & --- & --- \\
A2 & $1000$ real photographs & --- & --- \\
B & $1000$ renders drawn uniformly at random & dispersed & none \\
C10 & the $1000$ the allocation selects & concentrated & as scored \\
\midrule
CSHUF & C10's quotas, cluster scores permuted & fixed, same shape & destroyed \\
CINV & C10's quotas, their rank reversed & fixed, same shape & reversed \\
CORACLE & C10's quotas, reassigned by true endpoint error & fixed, same shape & true error \\
\midrule
A0FX, BFX, C10FX & A0, B, C10 with the optimisation budget unpinned & as A0, B, C10 & as A0, B, C10 \\
MT & nothing; A0's pool run single-round & --- & --- \\
\bottomrule
\end{tabular}
\end{center}
\end{table}

All arms (Table~\ref{tab:arms}) share the same base pool and differ
only in which $1000$ images are appended to it, so any difference between two of them is
attributable to that choice and nothing else. \textit{A0} appends
nothing. \textit{A2} appends $1000$ real photographs, which separates the effect of a larger pool
from the effect of generated content. \textit{B} appends $1000$ renders drawn uniformly at random,
redrawn per seed. \textit{C10} appends the $1000$ that the allocation above selects. A2, B and C10
all hold $5447$ pairs, and the comparison this paper turns on is C10 against B.

\begin{figure}[t]
\begin{center}
\includegraphics[width=0.95\textwidth]{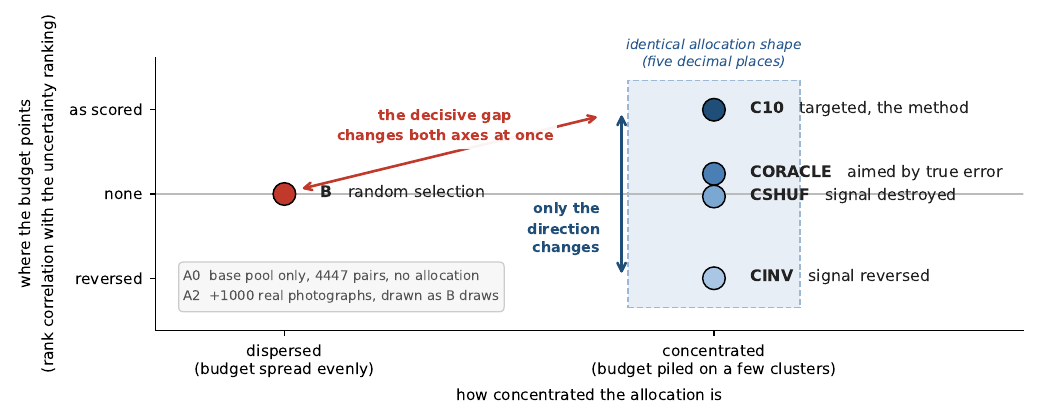}
\end{center}
\caption{\textit{What each arm holds fixed and what it varies.} The horizontal axis is how
concentrated an allocation is; the vertical axis is where it points, measured as rank correlation
with the committed uncertainty ranking. Because the C-family are permutations of one allocation,
they reproduce its shape to five decimal places and differ only vertically,  which is what makes
the vertical contrast clean and the diagonal one ambiguous. A0 and A2 append nothing and no
generated content, respectively, so neither has an allocation to place.
\emph{Sources:} \texttt{abc\_pools.json}; \texttt{T2\_RESULTS.md}; \texttt{results.md} \S5.2.}
\label{fig:arms}
\end{figure}

That comparison cannot settle the question on its own, and Figure~\ref{fig:arms} is why: a targeted
allocation differs from a random one along \emph{two} axes at once, it is more concentrated, and
it points somewhere, so a difference between C10 and B is attributable to neither. Three further
arms hold concentration exactly fixed and vary only where the budget points. \textit{CSHUF} permutes
the cluster scores, destroying the signal while preserving the allocation's shape; \textit{CINV}
reverses their rank; \textit{CORACLE} reassigns the quotas based on each cluster's true endpoint error.

Write $\hat{\sigma}$ for the seed-to-seed wobble, the pooled within-arm standard deviation of $S_\alpha$, taken at the conventional $\alpha = 0.5$. A gap is a \textsc{real effect} only if it exceeds twice that wobble \emph{and} every seed agrees on
its sign; a smaller gap is \textsc{within noise}; one that clears the bar without agreeing in sign is
\textsc{inconclusive}, and is reported as such rather than resolved by adding seeds. The rule applies
per architecture, and a disagreement between them is reported as a disagreement. We report no
$p$-values at three seeds. Appendix~\ref{app:method} records the rule's provenance, the seeds, and
the gates each campaign passed before launching.

\section{Results}
\label{sec:results}
Four analyses follow, each measured independently of the others: whether the targeted allocation
beats a random one on trained accuracy, what makes a targeted set measurably different, regardless of
that null, the mechanism behind the null, and a separate finding about the benchmark used to read all
of the above.

\subsection{Does Targeting Beat Random}
\label{sec:null}
The loop is specified, so the first question is the one it was built to answer. \textit{Aiming the budget by uncertainty did not beat spending it at random.} On the primary endpoint at eight seeds
per arm, the gap between the targeted allocation and a random one is $+0.0024$, about a third of
the spread it would have had to clear, and every one of the sixteen gaps that campaign decides
returns the same verdict.

A null is only as good as the objections it survives, so the rest of this section is those
objections and what each one was answered with. Table~\ref{tab:trained} is the whole trained record: four campaigns that return nothing, and one that returns something, which is how we know the four
nulls mean anything at all.

\begin{table}[t]
\caption{Trained comparison on the primary architecture and endpoint. A gap must
exceed its own campaign's bar \emph{and} agree in sign across seeds to count. Four campaigns attack
the decisive gap from different directions and none of them moves it; the positive control, which
compares against a method we expected to be clearly worse, clears its bar more than threefold.
Per-cell tables for both architectures and both endpoints are in Appendix~\ref{app:diag}.
\emph{Sources:} \texttt{abc\_verdict.json} and its \texttt{or/}, \texttt{fx/}, \texttt{se/}
variants; \texttt{pc\_verdict.json}.}
\label{tab:trained}
\begin{center}
\small
\setlength{\tabcolsep}{4pt}
\begin{tabular}{@{}llcccccl@{}}
\toprule
& What it tests & $n$ & Gap & $\Delta$ & $2\hat{\sigma}$ & $\times$\,bar & Verdict \\
\midrule
ABC & targeting vs.\ random, as committed & $3$ & C10$-$B & $+0.0050$ & $0.0179$ & $0.28$ & \textsc{within noise} \\
SE & the same gap at eight seeds & $8$ & C10$-$B & $+0.0024$ & $0.0069$ & $0.34$ & \textsc{within noise} \\
OR & a perfect score in place of ours & $3$ & CORACLE$-$B & $+0.0008$ & $0.0061$ & $0.14$ & \textsc{within noise} \\
FX & the optimisation budget unpinned & $3$ & C10FX$-$BFX & $+0.0014$ & $0.0116$ & $0.12$ & \textsc{within noise} \\
\midrule
PC & \emph{can this design see anything?} & $3$ & C10$-$MT & $+0.0197$ & $0.0060$ & $\mathbf{3.31}$ & \textsc{detected} \\
\bottomrule
\end{tabular}
\end{center}
\end{table}

That is the plain form of what the seed expansion returned, and it is worth stating before the arithmetic, because the
arithmetic looks backward. Eight seeds per arm give a truer estimate of the seed-to-seed spread
than three did: the real spread is \emph{larger} than the optimistic three-seed guess, while the gap
itself halved, from $+0.0050$ to $+0.0024$. So the bar sits at $0.0069$ on $28$ degrees of freedom,
not the $0.0034$ the pre-registration projected: $\hat{\sigma}$ is a per-run spread and does not
shrink with seeds, so that projection was an error. The gap now sits inside even the tightest bar
measured on this cell, $0.0055$. Appendix~\ref{app:diag} gives the per-arm spreads.

Each of the following is a reason the null might be an artifact rather than a result, and each has a campaign attached rather than a concession.

\emph{Perhaps we chose a bad uncertainty estimator}. The oracle campaign replaces the score with each cluster's \emph{true} error on the endpoint, information no real method could have, and nothing moves. All eight of its gaps are within noise, and not one is even directionally stable.
The substitution was not a small one, either. Our score and the truth it stands in for agree at a
rank correlation of $+0.2372$, so we replaced the signal with something several times better aimed, and the trained result did not notice. This converts \emph{our score did not help} into the stronger \emph{no score of any quality helped here}.

\emph{Perhaps the added images were never really trained on.} They were not, under the committed
schedule: steps per epoch are set by the shorter of two data loaders, so appended images displace
base-pool exposure instead of adding to it. The FX campaign removes that pin, which raises steps per
epoch from $253$ to $341$ on one architecture and $127$ to $171$ on the other. The decisive gap
stays within noise in all four cells and \emph{shrinks} to $+0.0014$. Growing the budget did not
reveal the effect the pin was hiding.

\emph{Perhaps three seeds are too few.} It was, and that is what the seed expansion above measures.
It is a separate campaign with its own arms, noise estimate, and verdicts rather than extra seeds
bolted onto a finished one, because adding seeds after seeing a result is how a null gets talked
into existence.

\emph{The instrument is not blind.} A null is only worth reading if the same measurement can
detect something real, so we trained an arm we expected to be clearly worse, a mean-teacher
baseline, same architecture, same seeds, same scorer, same rule, and asked whether this design
would see the difference. It did, comfortably and with every seed agreeing. And that control does not close the gap that matters most. It shows an effect of roughly the reference magnitude is visible; it does not show that an effect the size of our nulls would be.
The control's own bar, $0.0060$, is itself larger than the decisive gap it is meant to reassure us
about, so a positive control at $0.02$ licenses \emph{the instrument works}, not \emph{it would have
seen the effect we are nulling}.

The decisive statement is the pre-registered one above: no effect on the primary cell larger than the bar the seed expansion
measured. A post-hoc interval on the two arms actually compared is tighter still,
placing the gap in $[-0.0020, +0.0067]$, which \textit{excludes the reference effect by more than a
factor of two} while admitting effects a few times smaller. That interval decides nothing, the
frozen rule admits no interval but it is a stronger honest statement than ``we could not have
seen it''. Section~\ref{sec:concentration} reports the eight-seed equivalence tests.

\subsection{Concentration, Not Targeting}
\label{sec:concentration}
The null hypothesis states that the direction of the signal did not move the endpoint. It does not say that targeting
changes nothing, and it plainly does change something, because a targeted training set is easy to
tell apart from a random one. \textit{What makes it different is that the budget is concentrated on
a few clusters, not that it is aimed at the right ones.} The geometry of the selected set shows it;
trained accuracy at eight seeds is consistent with it, and the one family of metrics on which
concentration is large enough to resolve supports it, post hoc and on one architecture.

\begin{figure}[t]
\begin{center}
\includegraphics[width=\textwidth]{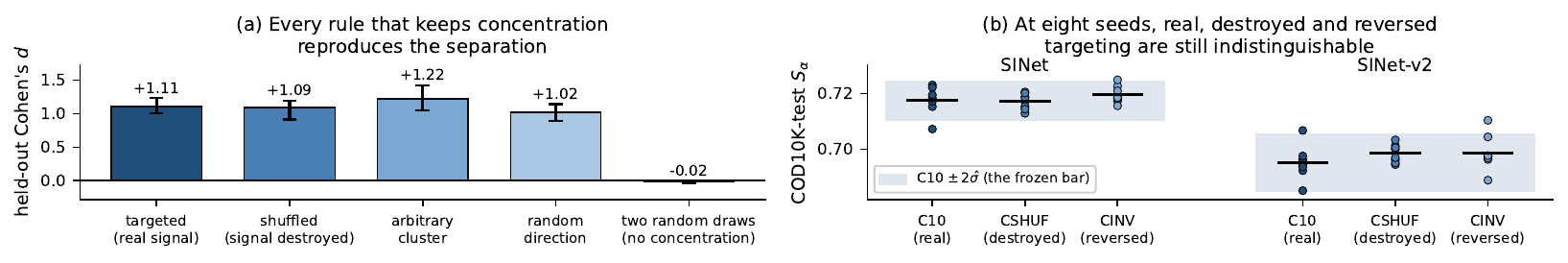}
\end{center}
\caption{\textbf{(a)} How far a selected set sits from a random set of the same size, in held-out
Cohen's $d$. Every rule that keeps the allocation's \emph{concentration} reproduces the separation;  only destroying concentration collapses it. Bars are means across the four cells at a budget of $N = 250$; whiskers show the range.
\textbf{(b)} Trained $S_\alpha$ at eight seeds for the three arms that hold concentration fixed and
vary only direction; points are seeds, rules are arm means, the band is the $\pm 2\hat{\sigma}$ bar.
\emph{Sources:} \texttt{c1\_attribution.csv}; \texttt{out/se/abc\_metrics.csv}.}
\label{fig:thesis}
\end{figure}

Aim a generation budget at the clusters where the model is least confident and the training set you get really does
look different from a randomly chosen one of the same size, train a probe to tell the two apart. It succeeds easily, on images it has never seen. However, re-run the selection with the uncertainty
scores shuffled between clusters, so the budget is spread exactly as unevenly while pointing at the
wrong places, and the separation survives. Aim at an arbitrary cluster instead: it survives again.
Only removing the unevenness itself, by drawing two random sets of equal size, collapses it (Figure~\ref{fig:thesis}a). The probe was detecting the concentration the whole time.

Paired across twenty cells of budget and configuration, the real signal contributes $+0.0073$ of a
Cohen's $d$ over its own shuffle, and against an arbitrary cluster it is \emph{negative}, at
$-0.0649$. On the strongest directional evidence we have, aiming does slightly worse than not
aiming. What targeting buys is narrowness rather than reach: its embedding covariance has a lower
effective rank than a random draw's in all twenty cells (mean ratio $0.634$), and covers no more of
the target manifold.

Geometry is not performance, so we trained it. The falsification arms hold the allocation's shape fixed to five decimal places and vary
only where the budget points, and at eight seeds per arm the arm that \emph{reverses} the
uncertainty signal, aiming the budget at the clusters the score calls easiest has the highest
mean score in \textit{four of the four} architecture--endpoint cells (Figure~\ref{fig:thesis}b).
This is not a claim that reversal helps, every one of those differences is within noise. It is
the absence of an ordering: if direction were what mattered, reversing it would cost something, and
nothing is paid. Even the destroyed-against-reversed contrast, which carries the strongest sign
agreement in the campaign at $8/8$ on one cell, does not reach two-thirds of its bar.

At eight seeds, a formal equivalence test becomes informative. However, it adds no asymmetry: five of
sixteen gaps are equivalent within $\pm 0.005$, four direction contrasts and one
targeting-versus-random gap, on NC4K (Table~\ref{tab:tost}). On the primary cell, targeting against
its own shuffle is equivalent ($p = 0.019$) and targeting against random is not ($p = 0.101$).
These tests are post-hoc and decide nothing.

The acquisition signal of Equation~\ref{eq:es} is boundary-focused, but $S_\alpha$ is structure-aware rather than boundary-specific, so the campaign never measured the quantity the signal was built to
move. Re-scoring every run's committed predictions with two boundary metrics closes that gap (Table~\ref{tab:boundary}), and on SINet, the concentration effect finally clears its bar. So does
the anti-targeted arm, at $1.34\times$ against targeted selection's $1.37\times$, the same effect
to within a tenth of a bar, while the two direction contrasts stay far inside it.
\textit{This is the strongest form of the result we have}, because it is the one place where the
same instrument resolves concentration and fails to resolve direction \emph{in the same cells}: the
null on direction is not simply an absence of sensitivity.

\begin{table}[t]
\caption{Arm differences on \emph{boundary} metrics, COD10K-test, same pooled-$\hat{\sigma}$ rule. \textbf{Post-hoc: not pre-registered, and no verdict is re-decided.} On SINet the concentration gap $\mathrm{C10}-\mathrm{B}$ clears its bar --- and so does the \emph{anti}-targeted $\mathrm{CINV}-\mathrm{B}$, by as much --- while both direction contrasts stay within noise. SINet-v2 resolves none of it. \emph{Source:} \texttt{diag\_subgroups.json}.}
\label{tab:boundary}
\begin{center}
\footnotesize
\setlength{\tabcolsep}{3.5pt}
\begin{tabular}{@{}llccccc@{}}
\toprule
Arch. & Metric & $2\hat{\sigma}$ & $\mathrm{C10}\!-\!\mathrm{B}$ & $\mathrm{CINV}\!-\!\mathrm{B}$ & $\mathrm{C10}\!-\!\mathrm{CSHUF}$ & $\mathrm{C10}\!-\!\mathrm{CINV}$ \\
\midrule
SINet & Boundary IoU & $0.0101$ & \textbf{$+0.0138$ 3/3} & \textbf{$+0.0135$ 3/3} & $+0.0038$ 2/3 & $+0.0004$ 2/3 \\
SINet & Boundary $F$ & $0.0151$ & \textbf{$+0.0190$ 3/3} & \textbf{$+0.0173$ 3/3} & $+0.0045$ 2/3 & $+0.0017$ 2/3 \\
SINet & IoU & $0.0109$ & \textbf{$+0.0112$ 3/3} & \textbf{$+0.0127$ 3/3} & $+0.0019$ 2/3 & $-0.0016$ 2/3 \\
\midrule
SINet-v2 & Boundary IoU & $0.0062$ & $+0.0010$ 2/3 & $+0.0025$ 2/3 & $-0.0002$ 2/3 & $-0.0015$ 3/3 \\
SINet-v2 & Boundary $F$ & $0.0065$ & $+0.0026$ 2/3 & $+0.0050$ 3/3 & $-0.0002$ 2/3 & $-0.0025$ 3/3 \\
SINet-v2 & IoU & $0.0100$ & $-0.0003$ 2/3 & $+0.0006$ 2/3 & $-0.0013$ 2/3 & $-0.0010$ 2/3 \\
\bottomrule
\end{tabular}
\end{center}
\footnotesize Bold marks a gap exceeding $2\hat{\sigma}$ with $3/3$ sign consistency; counts after
each $\Delta$ are sign-consistency across the three seeds. The pattern is not an artifact of where
predictions are binarised --- Appendix~\ref{app:diag} gives the threshold sweep.
\end{table}

These metrics were not pre-registered and re-decide no verdict; Table~\ref{tab:trained} stands. They binarise predictions
where $S_\alpha$ does not, though the pattern survives every binarisation threshold we swept. Moreover, \textit{SINet-v2 resolves none of it}, which we report rather than leading with the architecture
that cooperates.

\subsection{Why It Fails}
\label{sec:why}
Concentration explains why targeting \emph{looks} the way it does. It does not explain why this loop
could never have delivered the thing it promised, and that has a mechanism, one chain, measured end-to-end before any of the training runs above were scored.

Under every embedding we tried, $k$-means separation peaks at a
silhouette of $0.1600$ and is lower in the other two spaces, far below any conventional threshold
for calling a partition real. So a cluster-wise policy is drawing lines that the data does not
support; \textit{and even where those lines are real, the error does not sit between them.} Only
about a fifth of the endpoint's structural-error variance is between-cluster at all, so four fifths
of what a targeted policy would have to fix is invisible to \emph{any} cluster-level allocation,
however well aimed.

CAMO images are $24.8\%$ of the target pool, and the allocation is not neutral between the two components: $51.5\%$
of the budget lands on them, while the two largest quotas together hold $7$ of the $2026$ images the
paper is scored on (Table~\ref{tab:clustercomp}). ``Allocate by uncertainty'' turns out to be, in substantial part, ``allocate
to CAMO,'' \textit{and the quantity doing the allocating is itself the wrong one.} The
uncertainty score tracks per-pixel error about twice as strongly as it tracks the structural error
the endpoint measures ($+0.8553$ against $+0.4276$), and under whole-image aggregation, that ordering
holds for three families of uncertainty estimators on both architectures, so it is a property of uncertainty-guided allocation
here rather than a quirk of our particular score. Partialling out object area, the most plausible
mechanical explanation, \emph{raises} the structural correlation in every row, the opposite of what the confound predicts. It paints backgrounds and copies objects: the render set is a bijection onto the base foregrounds, and not one
of $8885$ traced objects shows any sign of having been regenerated. The distribution it produces is
narrower than its own input, synthetic images cover $0.13$ to $0.54$ of the real target manifold
($k$-NN recall, Appendix~\ref{app:coverage}), while the generator's own source photographs, real
pictures of an entirely different genre, cover $0.71$ to $0.75$. A policy allocating over that pool is choosing among copies.

\textit{One pattern points the other way, and we report it.} Pool \emph{size} may matter where pool
\emph{content} does not: A0, which appends nothing, trails the other arms consistently, reaching a
\textsc{real effect} on the secondary endpoint and two \textsc{inconclusive} verdicts once the
schedule is unpinned. No single step clears its bar, and we claim nothing from it, but it points away
from our conclusion, which is why it is here. Separately, we withdrew one of our own candidate
causes: we suspected the generator's conditioning channel was too narrow to act on a detailed
instruction, read the released source, and found four routes carrying foreground information, with
the dominant one bypassing the bottleneck we had blamed; so the explanation was wrong and is
retired rather than left standing. Appendix~\ref{app:tables} gives the full clustering sweep, the
signal-generality table, and the area control.

\subsection{CHAMELEON Is Mostly Training Data}
\label{sec:contamination}
The causes above are about our loop. This one is about the benchmark everyone uses, and it holds
whether or not our loop worked. Fifty of CHAMELEON's $76$ images, $65.8\%$ are the same photographs as images in the
COD10K-train and CAMO training pools, re-encoded, rescaled, or cropped. Forty-four have their
partner in the public COD10K-train split. A model trained on that split has therefore already seen
most of CHAMELEON before being evaluated on it. The standard check does not notice: hashing the
files returns \emph{zero} overlap, and it is not wrong, the copies were re-saved in a different
format, so not one byte matches while every pixel does. Sixteen images resolve neither way and are
reported \textsc{unchecked} rather than clean, so $50$ is a floor rather than an estimate.

\begin{figure}[t]
\begin{center}
\includegraphics[width=\textwidth]{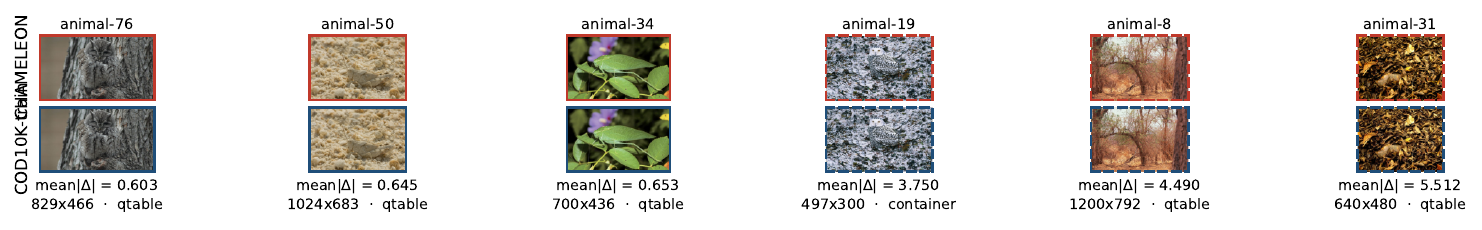}\\[0.45em]
\includegraphics[width=\textwidth]{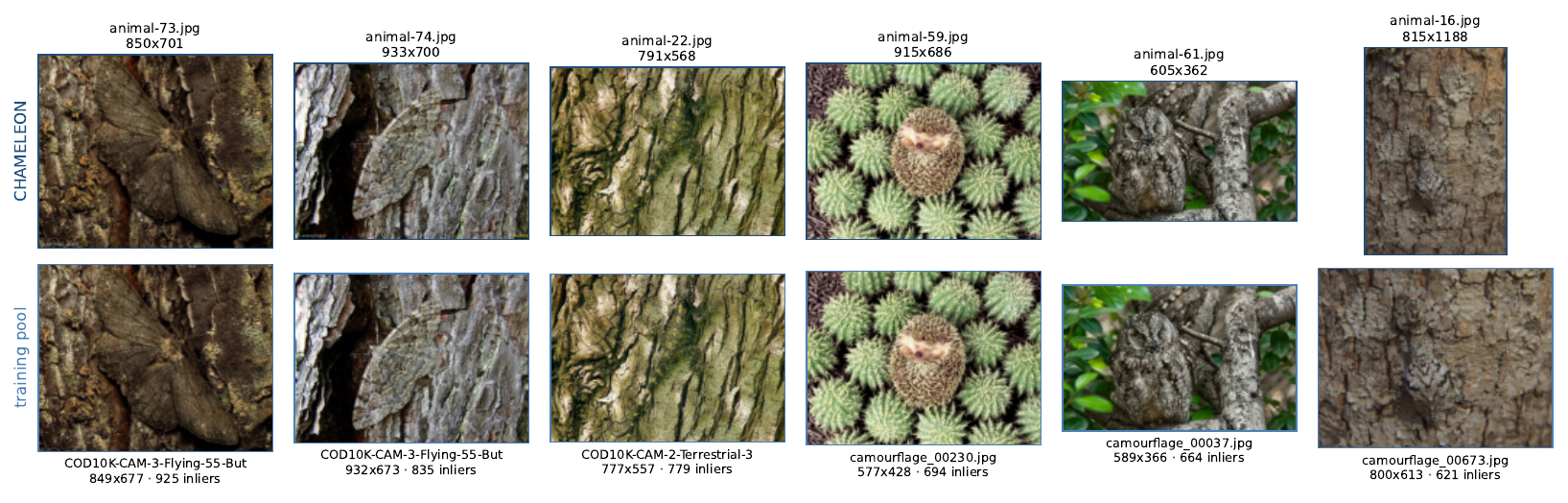}
\end{center}
\caption{Two tiers of the same finding. \textit{Top:} six of the $41$ same-dimension
re-encodes the CHAMELEON image above, its training-pool partner below. \textit{Bottom:} six of
the $9$ recovered by geometric adjudication, which the same-dimension detector reports
\textsc{unchecked} because a rescale or a crop moves an image out of its dimension group; note the
differing pixel dimensions in each pair's labels.}
\label{fig:cham}
\end{figure}

The count is measured in two tiers (Figure~\ref{fig:cham}). The floor is what a released
same-dimension detector reproduces on any machine: $41$ images whose partner shares their exact
pixel dimensions and differs only by re-encoding. A rescaled or cropped copy leaves its dimension
group and is invisible to that detector, so we recover those by matching keypoints under a geometric
model and then because agreeing on a homography is not the same as being one photograph, warping each candidate into the CHAMELEON frame and measuring the residual. \textit{Nine survive that test}; one cleared the keypoint threshold, failed the residual, and is excluded. Thresholds are calibrated on pairs whose answer is already committed, the finding is verified against a
byte-identical author-sourced release, and the same detector pointed at NC4K flags $0$ of $120$ (Appendix~\ref{app:chamext}).

We expected the contamination to flatter CHAMELEON scores, measured it, and did not find it: all eight difficulty percentiles for the leaked subset fall between $0.448$ and $0.559$. A protocol violation does not need a score advantage to
disqualify a benchmark. Nor do we recommend reporting CHAMELEON on its uncontaminated subset,
because that subset is now $10$ images, one of which is byte-identical to an
image in the split we select checkpoints on. \textit{A ten-image endpoint is not an endpoint:
report COD10K-test and NC4K instead, and disclose the unchecked share.}\footnote{This is not
hypothetical for the protocol we build on, which designates the real unlabelled COD10K training set
as its target domain. Testing on CHAMELEON, it removes those images from the \emph{source} domain, closing that path but not the target pool, which holds the same $44$. We attribute no intent: the source-side removal shows the authors were alert to leakage, and the target-side path is invisible without a re-encode-tolerant check.} The detector, the pair list, and the clean protocol
are released and run without any of our data.

\section{Discussion}
\label{sec:discussion}
Two kinds of evidence sit underneath this paper and should not be run together. The
\emph{geometric} result is positive, and we state it without hedging: concentration, not the
direction of the uncertainty signal, produces the separation that makes targeted selection look like it works. The \emph{downstream} result is a null under stated sensitivity, it does not show that uncertainty guidance cannot work, and we have named the size at which it would have. Four things transfer beyond this setup. \textit{Three checks cost no training and should precede any
cluster-wise allocation:} whether the data separates into clusters at all, what the funded clusters
are made of relative to the pool they are drawn from, and how much of the endpoint's error sits
between clusters rather than within them. \textit{The discriminating control is not a random
baseline} but a same-shape allocation with the signal destroyed inside it, because random selection
differs from targeting in two ways at once. \textit{A null on the endpoint metric does not license a
null on the quantity the acquisition signal was built to move.} And \textit{a real-versus-synthetic AUC near $1.0$ is near-vacuous evidence} ,  a lossy re-save of the identical images separates at
$0.9928$ (Appendix~\ref{app:coverage}).

The idea still seems reasonable. For it to have a chance, you would need an optimization budget that
grows with the pool rather than one under which added images displace real ones, a foreground supply
not already exhausted, a generator wider than its own input, and a target pool that is not a mixture
whose components the signal can tell apart. Nothing here says closed-loop generation cannot work,
only that this build could not have shown that it does. One pattern runs through the method and the
benchmark alike: \textit{a control that returns zero is informative only if it could have returned
something else.}

\section{Conclusion}
\label{sec:conclusion}
Under a fixed synthetic-data budget, aiming generation at the clusters a model finds hardest did not beat spending it at random. Across $103$ trained runs and five controls that each close a different
route to an artifactual null, the gap stays within noise. Targeting does change the training set,
but through concentration rather than aim, which only a same-shape control exposes. Independently,
$50$ of CHAMELEON's $76$ images are altered copies of COD10K and CAMO training images, which exact
hashing misses. Both findings argue for one practice: before crediting an acquisition rule or
trusting a duplicate check, run the version of the test that could have failed.

\bibliographystyle{plainnat}
\bibliography{reference}

\appendix

\section{Additional Related Work}
\label{app:related}

\paragraph{COD benchmarks and detectors.}
The standard camouflaged object detection (COD) evaluation suite comprises COD10K~\cite{fan2022concealed}, a large benchmark
with ten camouflage categories and pixel-accurate masks; CAMO~\cite{le2019anabranch}, which focuses
on man-made camouflage and occlusion; NC4K~\cite{lv2021simultaneously}, received from verified source; and
CHAMELEON~\cite{skurowski2018animal}, an earlier unpublished collection that has become a de facto
standard. The continued use of all four matters for Section~\ref{sec:contamination}: an image shared by a training set and any reported test set contaminates the evaluation for methods using the full suite.
SINet~\cite{fan2020sinet} decomposes COD into search and identification branches, while SINet-v2
\cite{fan2022concealed} adds a neighbour connection decoder for boundary refinement; we use both
off-the-shelf. SegMaR~\cite{jia2022segment} iteratively magnifies and re-processes ambiguous
regions, representing architectural progress outside our allocation experiment. We report the
structure-aware S-measure~\cite{fan2017structure}, enhanced-alignment E-measure
\cite{fan2018enhanced}, and weighted $F_\beta^w$~\cite{margolin2014evaluate} for comparability
with S2R-COD.

\paragraph{Synthetic generation and domain adaptation.}
Latent diffusion~\cite{rombach2022high} and DDIM~\cite{song2021denoising} provide the high-resolution
generator and a deterministic sampling schedule inherited by our pipeline. Copy-paste augmentation
\cite{ghiasi2021simple} similarly composes object cutouts with varied backgrounds; LAKE-RED applies
this compositional idea while retrieving background patches. CyCADA~\cite{hoffman2018cycada}, in
contrast, aligns source and target feature distributions with cycle-consistent adversarial training; our loop does not operate at the feature level and instead feeds generator output into a
self-training objective. Other camouflage-image generators exist~\cite{chen2025realcamo}, but they
are outside the experiment, so the reported null is bound to LAKE-RED.

\paragraph{Uncertainty-guided acquisition and self-training.}
FixMatch~\cite{sohn2020fixmatch} couples confidence thresholding with consistency regularisation; our pseudo-labels likewise arise in an inter-round consistency step, but drive a second training
round rather than an online update. Pseudo-labelling~\cite{lee2013pseudo} supplies the hard-target
view of this step. DINOv2~\cite{oquab2024dinov2} and CLIP~\cite{radford2021learning} provide the
frozen representations used for clustering. These choices complement core-set selection
\cite{sener2018active}: the clusters spread the budget, while the disagreement scores determine
which clusters receive more of it. BADGE~\cite{ash2020deep} jointly uses uncertainty and diversity
in gradient-embedding space; our same-shape shuffle in Section~\ref{sec:concentration} is a direct
control for the corresponding entanglement in this allocation.

\paragraph{Benchmark integrity and negative results.}
Dataset-collection procedures can induce shared statistical bias~\cite{torralba2011unbiased}, and
near-duplicate analyses of CIFAR show that visually identical images can cross split
boundaries~\cite{barz2020train}. Related contamination concerns appear in language-model evaluation
\cite{wu2024antileak}; the mechanism differs here, but the need for explicit auditing is shared.
Our pre-registration follows reproducibility guidance~\cite{pineau2021improving} and the case for
publishing disconfirming results~\cite{karl2024position}. Finally, criticisms of learned comparison
measures under resizing and representation choices~\cite{parmar2022aliased,jayasumana2024rethinking}
motivate treating encoding-sensitive probes as diagnostics rather than as clear perceptual evidence.

\paragraph{Where this work sits among budget-steering systems.} Table~\ref{tab:closest} places the
three systems discussed in Section~\ref{sec:related} beside this one. The last column is the point: it records whether a signal-destroying or same-shape control is run, which is what separates an
effect of \emph{aiming} a budget from an effect of \emph{concentrating} it. It is not a claim of
superiority, our own entry records a null.

\begin{table}[t]
\caption{Where this work sits among systems that steer a synthetic-data budget. \emph{Acquisition
signal} is what decides where the budget goes; \emph{generator control} is what the method can
actually vary in the generated image; \emph{diversity control} is what stops the budget piling onto
near-duplicates; \emph{closed-loop} means the downstream model's state feeds back into generation;
\emph{abstention/control} is whether a signal-destroying or same-shape control is run, which is what separates an effect of aiming from an effect of concentrating. The last column is the one this paper argues is missing from the literature, not a claim of superiority; our own entry records a
null.}
\label{tab:closest}
\begin{center}
\small
\setlength{\tabcolsep}{3pt}
\footnotesize
\begin{tabular}{@{}lp{1.75cm}p{1.9cm}p{1.75cm}p{1.35cm}p{1.9cm}@{}}
\toprule
Work & Task & Acquisition signal & Generator control & Closed-loop & Abstention / control \\
\midrule
LAKE-RED~\cite{zhao2024lakered} & COD synthesis & none (uniform) & background only & no & --- \\
S2R-COD~\cite{luo2025s2rcod} & COD adaptation & none (uniform) & background only & pseudo-labels & --- \\
GAUDA~\cite{frisch2025gauda} & surgical seg. & epistemic, per class & image + mask & yes, online & none reported \\
DisCL~\cite{liang2025discl} & long-tail cls. & hardness, per sample & guidance strength & curriculum & none reported \\
SynQuE~\cite{chen2025synque} & dataset ranking & proxy score, per set & n/a (picks sets) & no & none reported \\
\midrule
\textbf{This work} & COD adaptation & disagreement, per cluster & background only & pseudo-labels &
\textbf{same-shape shuffle, rank reversal, oracle} \\
\bottomrule
\end{tabular}
\end{center}
\end{table}

\section{Detailed Method Specification}
\label{app:method}

\paragraph{Training.} SINet~\cite{fan2020sinet} at batch $16$ for $40$ epochs and
SINet-v2~\cite{fan2022concealed} at batch $32$ for $100$ epochs, Adam at $10^{-4}$, images at
$352 \times 352$ with ImageNet normalisation. The exponential-moving-average (EMA) teacher uses momentum $0.996$ under this task
setting. The consistency weights in Equation~\ref{eq:es} are the configured $a = 0.9$, $b = 0.3$
with the unweighted cross-entropy variant, these are the values the trainer sets for this task,
not the loss class's own defaults. ES abbreviates the edge-aware saliency loss of the protocol we
build on; both of its terms are per-pixel means (the $\ell_1$ norm is averaged over pixels), and BCE
takes the teacher's output as its target. The acquisition score uses exactly this unweighted form; the training consistency loss uses the same two terms with a saliency-weighted BCE (per-pixel weight
equal to the teacher's output plus $c = 0.5$). Strong views add random autocontrast and Gaussian blur to the
weak view. \texttt{cudnn.deterministic} is enabled and \texttt{cudnn.benchmark} disabled, which
reduces, and is not claimed to eliminate, kernel nondeterminism.

\paragraph{Seeds and what was held constant.} The committed seeds are $\{42, 43, 45\}$; the seed
expansion of Section~\ref{sec:null} adds $\{46, 47, 48, 49, 50\}$, fixed in writing before its first
run. Across the four-arm campaign's $24$ runs, the base pool is asserted byte-identical against a
hash manifest with $0$ mismatches, both rounds read the unfiltered $4040$-image target set, and every run shares one trainer invocation, checkpoint rule, endpoint set, and deterministic-kernel
setting.

\paragraph{Optimization budget.} Steps per epoch are set by the shorter of the source and target
loaders, which pins them at $253$ (SINet) and $127$ (SINet-v2) in both rounds of every committed run
regardless of pool size, so appended images displace base-pool exposure rather than adding to it.
This is measured per run and asserted rather than assumed. It is not an asymmetry between the arms
that matter: B and C10 hold identical pool sizes and therefore receive identical step counts;
and Section~\ref{sec:null} reports the campaign that removes the pin and re-runs the comparison
without it.

\paragraph{Constructing the falsification arms.} CSHUF's permutation is the \textbf{first} of $64$
pre-declared draws satisfying zero fixed points and $|\rho| \leq 0.10$, chosen by a rule fixed
before the draw and never by inspecting its consequences; its realized rank correlation with the
true score vector is $-0.03351$, against CINV's exact $-1.0$. CINV's single fixed point is
mathematically necessary: at odd $k$ the median-ranked cluster is its own mirror. Because equality
is the \emph{supporting} outcome for these arms, two near-identical arms would manufacture it, so
set distinctness was gated before any GPU time at a Jaccard bound of $0.50$: between $452$ and $511$
of each pair's $1000$ injected images differ, at $2.2$--$2.4\times$ chance overlap.

\paragraph{Allocation rule.} Write $e_j$ for cluster $j$'s score (the mean of Equation~\ref{eq:es}
over its target members), $s_e$ for the population standard deviation of $e_1, \dots, e_k$, and $N$
for the generation budget. Cluster $j$ receives the share
\begin{equation}
p_j \;=\; \frac{\exp\!\big((e_j - \max_i e_i)/(\atemp\, s_e)\big)}{\sum_{l=1}^{k} \exp\!\big((e_l - \max_i e_i)/(\atemp\, s_e)\big)},
\label{eq:alloc}
\end{equation}
and the integer quotas are $\lfloor N p_j \rfloor$ plus one unit to each of the $N - \sum_j \lfloor N p_j \rfloor$
clusters with the largest remainders (ties to the lower index). Dividing by $s_e$ makes $\atemp$
scale-free: on the committed scores $s_e = 0.0101$, and Equation~\ref{eq:alloc} at $\atemp = 1.0$,
$N = 1000$ reproduces all $75$ committed quotas exactly (the ten largest are in
Table~\ref{tab:clustercomp}).

\paragraph{Allocation sweeps.} Stage (iv) matches against a grey-masked cutout rather than a render,
because at selection time, the foreground exists and the render does not. The budget grid is
$N \in \{250, 500, 1000, 2000, 3000, 4447\}$ and the temperature grid over $\atemp$ is extended
until both degenerate ends are reached, concentrated when $\max_j p_j \geq 0.95$, uniform when
total variation from uniform falls below $0.01$ with caps at $\atemp \in [10^{-4}, 10^{3}]$. The
campaign uses $\atemp = 1.0$, $N = 1000$.

\paragraph{Metrics.} $S_\alpha$ at $\alpha = 0.5$ is the primary endpoint metric; mean absolute
error (MAE) on CAMO-250 is the checkpoint-selection metric; $F_\beta^w$ and the E-measure $E_\phi$
are computed and reported, but they decide nothing.
The $F$-measures apply $\beta$ in the non-squared form, so the metric name rather than a formula is what identifies them.

\paragraph{Degrees of freedom.} Each campaign's $\hat{\sigma}$ is pooled within-arm across that
campaign's own arms, at $\mathrm{df} = (\text{arms}) \times (\text{seeds} - 1)$: $8$ for the
three-seed campaigns over four arms, $6$ for the three-arm controls, $4$ for the two-arm positive
control (C10 and MT), and $28$ for the eight-seed expansion. A campaign's bar is never swapped for another campaign's in either direction.

\paragraph{Campaign codes and verdicts.} Table~\ref{tab:trained} labels campaigns ABC (the committed
four-arm campaign: A0, A2, B, C10), SE (seed expansion), OR (oracle, arm CORACLE), FX (unpinned
optimization schedule, arms A0FX, BFX, C10FX), and PC (positive control against a mean-teacher arm,
MT); T2 is the falsification campaign (CSHUF, CINV). Every campaign but PC uses the three bands of
Section~\ref{sec:method}. PC's pre-registration names its outcomes \textsc{detected} ($\Delta > 2\hat{\sigma}$
with all seeds agreeing in sign) and \textsc{not detected} ($|\Delta| \le 2\hat{\sigma}$).

\paragraph{A2 controls one confound.} A2 pairs the authors' original photographs with the
authors' masks, whose mean foreground fraction differs from the raw masks by $0.0058$ on $1000$ of
$5447$ images. This affects $\Delta(\mathrm{B}-\mathrm{A2})$ and is disclosed; it does not touch
$\Delta(\mathrm{C10}-\mathrm{B})$, the decisive gap, since B and C10 share mask provenance, pool
size and render pool.

\paragraph{Pre-registration commits and amendments.} Every decision rule was committed before the
runs it governs. The four-arm rule is recorded at \texttt{065dac6}, the commit carrying that
campaign's first log block; the falsification campaign's at \texttt{617a2e1}; the signal-generality
check's at \texttt{6efde9f}; the area control's and the positive control's at \texttt{c2114af}; the
seed expansion's at \texttt{b755aa9}, dated 2026-09-18. The oracle and optimization schedule
campaigns carry their own pre-registrations, each committed before that campaign's first training
run. Three amendments exist, all append-only and dated. Two are reference-tolerance defects of the
same kind: a frozen clause requiring reproduction to better than $10^{-9}$ against a file recording
six decimal places, and in the area control, written by the same author who had just documented
the first, the identical mistake against a file recording four. Both were replaced by exact
agreement at the recorded precision, which is the strictest satisfiable form, and neither altered a
decision threshold. The third marks one verification gate not applicable in scope after it was found
to compare two architectures' signals.

\section{Decisive Gap in Every Cell, and the Bars It Is Read Against}
\label{app:diag}

Table~\ref{tab:trained} reports the primary cell. Table~\ref{tab:allcells} reports all four for
each campaign that attacks the decisive gap. \textbf{Sixteen of sixteen are \textsc{within noise},
and not one reaches half of its bar.}

\begin{table}[h]
\caption{The decisive gap in all four architectures-endpoint cells, per campaign. Each campaign is
read against its \emph{own} pre-registered bar, never another's. \emph{Source:}
\texttt{abc\_verdict.json} and its \texttt{se/}, \texttt{or/}, \texttt{fx/} variants.}
\label{tab:allcells}
\begin{center}
\small
\begin{tabular}{@{}llcccccc@{}}
\toprule
& Cell & $n$ & Gap & $\Delta$ & $2\hat{\sigma}$ & $\times$\,bar & sign \\
\midrule
ABC & SINet $\cdot$ COD10K & 3 & C10$-$B & $+0.0050$ & $0.0179$ & $0.28$ & 3/3 \\
 & SINet $\cdot$ NC4K &  &  & $+0.0021$ & $0.0083$ & $0.25$ & 3/3 \\
 & SINet-v2 $\cdot$ COD10K &  &  & $-0.0004$ & $0.0123$ & $0.03$ & 2/3 \\
 & SINet-v2 $\cdot$ NC4K &  &  & $+0.0009$ & $0.0105$ & $0.09$ & 2/3 \\
\midrule
SE & SINet $\cdot$ COD10K & 8 & C10$-$B & $+0.0024$ & $0.0069$ & $0.34$ & 6/8 \\
 & SINet $\cdot$ NC4K &  &  & $+0.0003$ & $0.0059$ & $0.06$ & 4/8 \\
 & SINet-v2 $\cdot$ COD10K &  &  & $-0.0011$ & $0.0101$ & $0.11$ & 5/8 \\
 & SINet-v2 $\cdot$ NC4K &  &  & $-0.0009$ & $0.0093$ & $0.10$ & 4/8 \\
\midrule
OR & SINet $\cdot$ COD10K & 3 & CORACLE$-$B & $+0.0008$ & $0.0061$ & $0.14$ & 2/3 \\
 & SINet $\cdot$ NC4K &  &  & $-0.0012$ & $0.0047$ & $0.26$ & 2/3 \\
 & SINet-v2 $\cdot$ COD10K &  &  & $+0.0023$ & $0.0093$ & $0.25$ & 2/3 \\
 & SINet-v2 $\cdot$ NC4K &  &  & $+0.0020$ & $0.0080$ & $0.25$ & 2/3 \\
\midrule
FX & SINet $\cdot$ COD10K & 3 & C10FX$-$BFX & $+0.0014$ & $0.0116$ & $0.12$ & 2/3 \\
 & SINet $\cdot$ NC4K &  &  & $+0.0023$ & $0.0107$ & $0.21$ & 2/3 \\
 & SINet-v2 $\cdot$ COD10K &  &  & $-0.0007$ & $0.0100$ & $0.07$ & 1/3 \\
 & SINet-v2 $\cdot$ NC4K &  &  & $-0.0003$ & $0.0122$ & $0.03$ & 1/3 \\
\bottomrule
\end{tabular}
\end{center}
\end{table}

\paragraph{Why the committed bar is so wide, and why we did not narrow it.} The four-arm campaign
pools its within-arm spread across all four of its arms, and one of them dominates. On the primary
cell the per-arm standard deviations are A0 $0.01727$, A2 $0.00193$, B $0.00181$ and C10 $0.00406$:
\textbf{A0 alone contributes $92.7\%$ of the pooled variance}, and A0 appears in the decisive
comparison not at all. Its instability is unexplained and is reported rather than excluded, because
the bar was pre-registered as a pooled quantity, and re-deriving it after seeing the result is
precisely the move the pre-registration exists to prevent. What we do instead is report the tighter
bars that later campaigns measured on their own arms, and let the reader see all of them.

\paragraph{Three bars, and what each is good for.} Three estimates of the same wobble exist,
and they are not interchangeable. The committed four-arm bar is $0.0179$ and is inflated by A0 as
above. The falsification campaign, pooling over B, C10, CSHUF, and CINV; the arms that actually appear in a decisive gap, yields a measure of $0.0055$ at the same degrees of freedom. The seed expansion
measures $0.0069$ over the same four arms at $28$ degrees of freedom instead of $8$. The asymmetry
that matters is this: where equality is the \emph{supporting} outcome, a tighter bar is
conservative, but here the claim and the null coincide, so a looser bar flatters us. That is why the
paper reads the gap against the tightest defensible bar rather than the widest, and why the seed
expansion's bar, not the committed one, is the sensitivity statement in Section~\ref{sec:null}.

\paragraph{Bar widened when the seeds grew, and the projection it replaced was wrong.} The seed
expansion's pre-registration projected that eight seeds would tighten the bar by
$\sqrt{8/3} \approx 1.63\times$ if the per-arm spreads stayed as measured. That projection was an
error in the pre-registration: $\hat{\sigma}$ is a pooled \emph{per-run} standard deviation, whose
expected value does not depend on the number of seeds, so at unchanged spreads the bar would have
stayed at $0.0055$; a $\sqrt{8/3}$ reduction applies to the standard error of a difference of means,
which the verdict rule does not use. The spreads did not stay put, either. Everyone grew: on the primary cell, B went $0.00181 \rightarrow 0.00251$, C10 $0.00406 \rightarrow 0.00489$, CSHUF
$0.00166 \rightarrow 0.00278$ and CINV $0.00285 \rightarrow 0.00302$. A pooled standard deviation
on eight degrees of freedom is a noisy estimate that lands low as often as high; on twenty-eight, it
is far steadier. So the expansion could never have bought a smaller bar; it bought a trustworthy
one, and showed the three-seed bars to have been optimistic rather than conservative.

\begin{table}[t]
\caption{Post-hoc interval estimates at \textbf{eight seeds per arm}, $S_\alpha$ on both endpoints.
\emph{These decide nothing}: verdicts remain the pre-registered $2\hat{\sigma}$ rule of
Section~\ref{sec:method}. Intervals are Welch on the two arms actually compared, so they are not
inflated by an arm outside the comparison. TOST (two one-sided tests;~\cite{lakens2017equivalence}) is run at $\delta = 0.005$; bold marks the
comparisons that reject, and the final column gives the smallest $\delta$ at which each comparison
\emph{could} have concluded equivalence. \emph{Source:} \texttt{out/se/abc\_metrics.csv}, via
\texttt{figs/make\_tost\_table.py} using \texttt{make\_results.py}'s \texttt{welch\_tost}.}
\label{tab:tost}
\begin{center}
\footnotesize
\setlength{\tabcolsep}{4pt}
\begin{tabular}{@{}lllcccc@{}}
\toprule
Arch. & Endpoint & Gap & $\Delta$ & 95\% CI & TOST $p$ & smallest $\delta$ \\
\midrule
SINet & COD10K & $\mathrm{C10}-\mathrm{B}$ & $+0.0024$ & $[-0.0020, +0.0067]$ & 0.1009 & $0.0059$ \\
SINet & COD10K & $\mathrm{C10}-\mathrm{CSHUF}$ & $+0.0003$ & $[-0.0041, +0.0047]$ & \textbf{0.0189} & $0.0039$ \\
SINet & COD10K & $\mathrm{C10}-\mathrm{CINV}$ & $-0.0021$ & $[-0.0066, +0.0023]$ & 0.0905 & $0.0057$ \\
SINet & COD10K & $\mathrm{CSHUF}-\mathrm{CINV}$ & $-0.0024$ & $[-0.0055, +0.0007]$ & \textbf{0.0486} & $0.0050$ \\
\midrule
SINet & NC4K & $\mathrm{C10}-\mathrm{B}$ & $+0.0003$ & $[-0.0033, +0.0040]$ & \textbf{0.0084} & $0.0033$ \\
SINet & NC4K & $\mathrm{C10}-\mathrm{CSHUF}$ & $+0.0007$ & $[-0.0029, +0.0042]$ & \textbf{0.0106} & $0.0036$ \\
SINet & NC4K & $\mathrm{C10}-\mathrm{CINV}$ & $-0.0031$ & $[-0.0067, +0.0006]$ & 0.1314 & $0.0060$ \\
SINet & NC4K & $\mathrm{CSHUF}-\mathrm{CINV}$ & $-0.0037$ & $[-0.0064, -0.0011]$ & 0.1622 & $0.0059$ \\
\midrule
SINet-v2 & COD10K & $\mathrm{C10}-\mathrm{B}$ & $-0.0011$ & $[-0.0067, +0.0045]$ & 0.0764 & $0.0057$ \\
SINet-v2 & COD10K & $\mathrm{C10}-\mathrm{CSHUF}$ & $-0.0033$ & $[-0.0087, +0.0020]$ & 0.2516 & $0.0077$ \\
SINet-v2 & COD10K & $\mathrm{C10}-\mathrm{CINV}$ & $-0.0035$ & $[-0.0101, +0.0031]$ & 0.3143 & $0.0089$ \\
SINet-v2 & COD10K & $\mathrm{CSHUF}-\mathrm{CINV}$ & $-0.0001$ & $[-0.0057, +0.0054]$ & \textbf{0.0402} & $0.0047$ \\
\midrule
SINet-v2 & NC4K & $\mathrm{C10}-\mathrm{B}$ & $-0.0009$ & $[-0.0068, +0.0049]$ & 0.0790 & $0.0057$ \\
SINet-v2 & NC4K & $\mathrm{C10}-\mathrm{CSHUF}$ & $-0.0023$ & $[-0.0078, +0.0031]$ & 0.1534 & $0.0068$ \\
SINet-v2 & NC4K & $\mathrm{C10}-\mathrm{CINV}$ & $-0.0043$ & $[-0.0102, +0.0016]$ & 0.3945 & $0.0091$ \\
SINet-v2 & NC4K & $\mathrm{CSHUF}-\mathrm{CINV}$ & $-0.0019$ & $[-0.0061, +0.0022]$ & 0.0655 & $0.0053$ \\
\bottomrule
\end{tabular}
\end{center}
\footnotesize At three seeds no comparison reached equivalence at $\delta = 0.005$. At eight,
five of sixteen do: four direction contrasts and the targeting-versus-random gap on SINet/NC4K. The
targeting-versus-random gap on the primary cell does not.
\end{table}

\paragraph{The interval at three seeds, and why it was left out of the main text.} A Welch interval on the two
arms actually compared, at the committed three seeds, places the decisive gap in
$[-0.0035, +0.0136]$ and excludes the reference effect. The eight-seed interval supersedes it
reported in Section~\ref{sec:null} and is recorded here only for continuity with the committed
account: at roughly $2.8$ degrees of freedom, its exclusion held by a margin too thin to lean on,
which is exactly the weakness the seed expansion removes.

\paragraph{Boundary metrics: the binarisation sweep in full.} Boundary metrics binarise a
prediction where $S_\alpha$ does not, so the result in Section~\ref{sec:concentration} could in principle be an artifact of where the threshold sits. It is not. Swept over thresholds
$0.4 / 0.5 / 0.6$ on SINet Boundary-IoU~\cite{cheng2021boundary}, $\Delta(\mathrm{C10}-\mathrm{B})$ holds at
$1.39 / 1.37 / 1.23 \times$ its bar and $\Delta(\mathrm{CINV}-\mathrm{B})$ at
$1.25 / 1.34 / 1.26 \times$, both $3/3$ sign-consistent at every threshold, while the two direction
contrasts never exceed $0.52\times$ and $0.14\times$ anywhere in the sweep. Concentration resolves
at every threshold; direction resolves at none.

\paragraph{How much the trained comparison is attenuated.} The trained half of the concentration
result is diluted three times over, and we disclose it because the attenuation biases toward the
null we report. Injected images are $1000$ of $5447$ pairs, or $18.4\%$ of the pool; under the
committed schedule, the pinned step count dilutes their exposure further; and only about $48\%$ of
injected images differ between any two arms, giving an effective contrast near $8.8\%$. Two of the
three are now measured rather than argued: the optimization-schedule campaign removes the pin and
the decisive gap \emph{shrinks}, and the seed expansion raises the degrees of freedom behind the
bar from $8$ to $28$. What remains is the third, which is a property of comparing two arms that
share their whole base pool and roughly half of their injected images by construction.

\section{Supplementary Tables}
\label{app:tables}

\begin{table}[h]
\caption{Full clustering sweep over the target set: silhouette (mean over 10 $k$-means seeds),
bootstrap ARI and seed ARI, for every $k$ in the declared grid and all three embedding spaces.
The principled $k$ per space is \textbf{bold}. No $k$ in any space reaches a silhouette conventionally
regarded as well-separated structure, and \texttt{clipL224} falls almost monotonically, so its selection
criterion returns the grid edge rather than an interior maximum --- recorded as a FAIL of the declared
guard. \emph{Source:} \texttt{b1\_k\_sweep\_*.csv};
block \texttt{EXP B1} \#3 @ \texttt{results/REBUILD\_LOG.txt:922}.}
\label{tab:silhouette}
\begin{center}\small
\begin{tabular}{rcccccccccc}
\toprule
& \multicolumn{3}{c}{\texttt{dinoL224}} & \multicolumn{3}{c}{\texttt{dinoL518}} & \multicolumn{3}{c}{\texttt{clipL224}} \\
\cmidrule(lr){2-4}\cmidrule(lr){5-7}\cmidrule(lr){8-10}
$k$ & sil. & bs.ARI & sd.ARI & sil. & bs.ARI & sd.ARI & sil. & bs.ARI & sd.ARI \\
\midrule
5 & 0.0538 & 0.712 & 0.795 & 0.0566 & 0.624 & 0.732 & \textbf{0.0568} & 0.953 & 0.983 \\
10 & 0.0868 & 0.676 & 0.709 & 0.0923 & 0.598 & 0.770 & 0.0550 & 0.692 & 0.805 \\
15 & 0.1053 & 0.608 & 0.714 & 0.1152 & 0.573 & 0.592 & 0.0486 & 0.575 & 0.670 \\
20 & 0.1180 & 0.629 & 0.693 & 0.1326 & 0.506 & 0.524 & 0.0486 & 0.635 & 0.646 \\
30 & 0.1378 & 0.596 & 0.642 & 0.1527 & 0.539 & 0.617 & 0.0488 & 0.609 & 0.631 \\
50 & \textbf{0.1465} & 0.610 & 0.615 & 0.1556 & 0.579 & 0.600 & 0.0448 & 0.498 & 0.549 \\
75 & 0.1457 & 0.588 & 0.627 & \textbf{0.1600} & 0.579 & 0.616 & 0.0396 & 0.463 & 0.477 \\
100 & 0.1419 & 0.608 & 0.637 & 0.1557 & 0.583 & 0.601 & 0.0382 & 0.419 & 0.452 \\
150 & 0.1367 & 0.630 & 0.636 & 0.1507 & 0.618 & 0.625 & 0.0357 & 0.380 & 0.409 \\
\bottomrule
\end{tabular}
\end{center}
\end{table}

\begin{table}[h]
\caption{Full signal-generality measurement, all 16 rows: four signal variants across three signal
families, two architectures, two aggregations. $\rho$ at the committed seed, $\pm$ sd over 10
$k$-means seeds. The whole-image block passes the pre-registered ordering $8/8$; the boundary block
passes $2/8$ with four rows going negative, an outcome outside the pre-registered interpretation
space, reported as it fell. \emph{Source:} \texttt{rebuild/T2C/out/t2c\_table.csv}; block
\texttt{EXP T2C} @ \texttt{results/REBUILD\_LOG.txt:2550}.}
\label{tab:t2cfull}
\begin{center}\footnotesize
\setlength{\tabcolsep}{1.5pt}
\begin{tabular}{llcccccc}
\toprule
arch & signal & agg. & $\rho(\mathrm{MAE})$ & $\rho(1-S_\alpha)$ & $\rho(1-\mathrm{IoU})$ & order & seeds \\
\midrule
SINet-v2 & ES & whole & $+0.7070 \pm 0.0404$ & $+0.5628 \pm 0.0359$ & $+0.4591 \pm 0.0315$ & PASS & 10/10 \\
SINet-v2 & ensemble A0 & whole & $+0.6191 \pm 0.0675$ & $+0.5483 \pm 0.0409$ & $+0.4540 \pm 0.0451$ & PASS & 9/10 \\
SINet-v2 & ensemble CSHUF & whole & $+0.5576 \pm 0.0690$ & $+0.4488 \pm 0.0580$ & $+0.3370 \pm 0.0577$ & PASS & 10/10 \\
SINet-v2 & entropy & whole & $+0.7393 \pm 0.0309$ & $+0.5404 \pm 0.0408$ & $+0.4070 \pm 0.0395$ & PASS & 10/10 \\
SINet-v2 & ES & bound & $-0.5485 \pm 0.0745$ & $-0.1146 \pm 0.0675$ & $+0.0537 \pm 0.0558$ & FAIL & 0/10 \\
SINet-v2 & ensemble A0 & bound & $-0.1960 \pm 0.0524$ & $+0.0425 \pm 0.0410$ & $+0.0933 \pm 0.0396$ & FAIL & 0/10 \\
SINet-v2 & ensemble CSHUF & bound & $-0.2527 \pm 0.0668$ & $-0.0591 \pm 0.0609$ & $-0.0011 \pm 0.0579$ & FAIL & 0/10 \\
SINet-v2 & entropy & bound & $+0.4208 \pm 0.0356$ & $+0.4182 \pm 0.0455$ & $+0.3605 \pm 0.0470$ & PASS & 8/10 \\
SINet & ES & whole & $+0.6284 \pm 0.0381$ & $+0.3433 \pm 0.0301$ & $+0.2613 \pm 0.0383$ & PASS & 10/10 \\
SINet & ensemble A0 & whole & $+0.6028 \pm 0.0466$ & $+0.2698 \pm 0.0487$ & $+0.1491 \pm 0.0560$ & PASS & 10/10 \\
SINet & ensemble CSHUF & whole & $+0.5212 \pm 0.0564$ & $+0.4006 \pm 0.0459$ & $+0.3361 \pm 0.0447$ & PASS & 10/10 \\
SINet & entropy & whole & $+0.5892 \pm 0.0368$ & $+0.4267 \pm 0.0277$ & $+0.3588 \pm 0.0351$ & PASS & 10/10 \\
SINet & ES & bound & $-0.0012 \pm 0.0481$ & $-0.0567 \pm 0.0500$ & $-0.0554 \pm 0.0479$ & FAIL & 8/10 \\
SINet & ensemble A0 & bound & $+0.3633 \pm 0.0605$ & $+0.2910 \pm 0.0509$ & $+0.2215 \pm 0.0513$ & PASS & 10/10 \\
SINet & ensemble CSHUF & bound & $+0.3175 \pm 0.0637$ & $+0.3857 \pm 0.0381$ & $+0.3770 \pm 0.0322$ & FAIL & 2/10 \\
SINet & entropy & bound & $+0.2434 \pm 0.0414$ & $+0.2492 \pm 0.0540$ & $+0.2432 \pm 0.0563$ & FAIL & 6/10 \\
\bottomrule
\end{tabular}
\end{center}
\end{table}

\begin{table}[h]
\caption{The area control (\texttt{EXP AC}). First-order partial Spearman on the same cluster
partition and floors as the committed measurement. $z_{\mathrm{obj}}$ = per-cluster mean object area
on the endpoint side; $z_{\mathrm{unc}}$ = per-cluster mean uncertain-region area on the target side.
Ordering survives $7/8$ and $6/8$ against a floor of $6/8$ declared before the run. Note that
partialling object area \emph{raises} $\rho(1-S_\alpha)$ in every row --- the opposite of the
confound's prediction. \emph{Source:} \texttt{rebuild/AC/out/ac\_table.csv}; block \texttt{EXP AC} @
\texttt{results/REBUILD\_LOG.txt:2612}.}
\label{tab:areacontrol}
\begin{center}\small
\begin{tabular}{llcccccc}
\toprule
 & & \multicolumn{2}{c}{raw} & \multicolumn{2}{c}{$|\,z_{\mathrm{obj}}$} & \multicolumn{2}{c}{$|\,z_{\mathrm{unc}}$} \\
\cmidrule(lr){3-4}\cmidrule(lr){5-6}\cmidrule(lr){7-8}
arch & signal & MAE & $1-S_\alpha$ & MAE & $1-S_\alpha$ & MAE & $1-S_\alpha$ \\
\midrule
SINet & ES & $+0.6284$ & $+0.3433$ & $+0.5761$ & $+0.4988$ & $+0.5518$ & $+0.3897$ \\
SINet & entropy & $+0.5892$ & $+0.4267$ & $+0.5505$ & $+0.5517$ & $+0.5060$ & $+0.4426$ \\
SINet & ensemble A0 & $+0.6028$ & $+0.2698$ & $+0.5411$ & $+0.4301$ & $+0.5603$ & $+0.2583$ \\
SINet & ensemble CSHUF & $+0.5212$ & $+0.4006$ & $+0.5002$ & $+0.4857$ & $+0.4857$ & $+0.3937$ \\
SINet-v2 & ES & $+0.7070$ & $+0.5628$ & $+0.7062$ & $+0.6272$ & $+0.4457$ & $+0.4744$ \\
SINet-v2 & entropy & $+0.7393$ & $+0.5404$ & $+0.7134$ & $+0.6546$ & $+0.5369$ & $+0.4266$ \\
SINet-v2 & ensemble A0 & $+0.6191$ & $+0.5483$ & $+0.6245$ & $+0.5969$ & $+0.4355$ & $+0.4488$ \\
SINet-v2 & ensemble CSHUF & $+0.5576$ & $+0.4488$ & $+0.5344$ & $+0.5209$ & $+0.4482$ & $+0.3615$ \\
\midrule
\multicolumn{2}{l}{ordering holds} & \multicolumn{2}{c}{$8/8$} & \multicolumn{2}{c}{$7/8$} & \multicolumn{2}{c}{$6/8$} \\
\bottomrule
\end{tabular}
\end{center}
\end{table}

\begin{table}[p]
\caption{Per-run metrics at full recorded precision, COD10K-test endpoint, for every run of the two
frozen campaigns and the mean-teacher positive control --- $39$ of the $103$ training runs in this
work; Table~\ref{tab:perrun-nc4k} gives the same runs on NC4K. Arms B and C10 appear once; they were
re-scored, not re-trained, for the falsification campaign and reproduced their committed values
exactly at six decimal places. \emph{Source:} \texttt{rebuild/ABC/out/abc\_metrics.csv},
\texttt{out/t2/abc\_metrics.csv} and \texttt{rebuild/PC/out/pc\_metrics.csv}; blocks
\texttt{EXP ABC} \#3 and \texttt{EXP T2} \#3.}
\label{tab:perrun}
\begin{center}\small
\begin{tabular}{llllcc}
\toprule
arch & arm & seed & endpoint & $S_\alpha$ & MAE \\
\midrule
SINet & A0 & 42 & COD10K & 0.715086 & 0.074960 \\
SINet & A0 & 43 & COD10K & 0.681706 & 0.077044 \\
SINet & A0 & 45 & COD10K & 0.706058 & 0.077347 \\
SINet & A2 & 42 & COD10K & 0.705507 & 0.075994 \\
SINet & A2 & 43 & COD10K & 0.708334 & 0.078489 \\
SINet & A2 & 45 & COD10K & 0.709196 & 0.080858 \\
SINet & B & 42 & COD10K & 0.711554 & 0.076085 \\
SINet & B & 43 & COD10K & 0.713632 & 0.076278 \\
SINet & B & 45 & COD10K & 0.715155 & 0.075917 \\
SINet & C10 & 42 & COD10K & 0.717165 & 0.073184 \\
SINet & C10 & 43 & COD10K & 0.723034 & 0.075136 \\
SINet & C10 & 45 & COD10K & 0.715243 & 0.078355 \\
SINet & CINV & 42 & COD10K & 0.722612 & 0.073910 \\
SINet & CINV & 43 & COD10K & 0.717343 & 0.077323 \\
SINet & CINV & 45 & COD10K & 0.718084 & 0.073607 \\
SINet & CSHUF & 42 & COD10K & 0.718602 & 0.074840 \\
SINet & CSHUF & 43 & COD10K & 0.716974 & 0.077678 \\
SINet & CSHUF & 45 & COD10K & 0.715291 & 0.079340 \\
SINet & MT & 42 & COD10K & 0.699633 & 0.093377 \\
SINet & MT & 43 & COD10K & 0.699167 & 0.087571 \\
SINet & MT & 45 & COD10K & 0.697503 & 0.091322 \\
SINetv2 & A0 & 42 & COD10K & 0.688813 & 0.091691 \\
SINetv2 & A0 & 43 & COD10K & 0.678740 & 0.095147 \\
SINetv2 & A0 & 45 & COD10K & 0.699883 & 0.085756 \\
SINetv2 & A2 & 42 & COD10K & 0.691914 & 0.089186 \\
SINetv2 & A2 & 43 & COD10K & 0.688977 & 0.094524 \\
SINetv2 & A2 & 45 & COD10K & 0.695414 & 0.087455 \\
SINetv2 & B & 42 & COD10K & 0.698340 & 0.092242 \\
SINetv2 & B & 43 & COD10K & 0.689481 & 0.089883 \\
SINetv2 & B & 45 & COD10K & 0.697385 & 0.087132 \\
SINetv2 & C10 & 42 & COD10K & 0.692321 & 0.093395 \\
SINetv2 & C10 & 43 & COD10K & 0.696328 & 0.090749 \\
SINetv2 & C10 & 45 & COD10K & 0.695360 & 0.089182 \\
SINetv2 & CINV & 42 & COD10K & 0.697008 & 0.088438 \\
SINetv2 & CINV & 43 & COD10K & 0.696331 & 0.087293 \\
SINetv2 & CINV & 45 & COD10K & 0.697605 & 0.090189 \\
SINetv2 & CSHUF & 42 & COD10K & 0.701269 & 0.087506 \\
SINetv2 & CSHUF & 43 & COD10K & 0.694490 & 0.087305 \\
SINetv2 & CSHUF & 45 & COD10K & 0.696218 & 0.089485 \\
\bottomrule
\end{tabular}
\end{center}
\end{table}

\begin{table}[p]
\caption{Per-run metrics at full recorded precision, NC4K endpoint, for the same $39$ runs as
Table~\ref{tab:perrun}. \emph{Source:} as Table~\ref{tab:perrun}.}
\label{tab:perrun-nc4k}
\begin{center}\small
\begin{tabular}{llllcc}
\toprule
arch & arm & seed & endpoint & $S_\alpha$ & MAE \\
\midrule
SINet & A0 & 42 & NC4K & 0.765121 & 0.085752 \\
SINet & A0 & 43 & NC4K & 0.751545 & 0.088874 \\
SINet & A0 & 45 & NC4K & 0.757700 & 0.090228 \\
SINet & A2 & 42 & NC4K & 0.760908 & 0.088385 \\
SINet & A2 & 43 & NC4K & 0.762933 & 0.087466 \\
SINet & A2 & 45 & NC4K & 0.765789 & 0.090892 \\
SINet & B & 42 & NC4K & 0.764108 & 0.087897 \\
SINet & B & 43 & NC4K & 0.768972 & 0.085001 \\
SINet & B & 45 & NC4K & 0.767632 & 0.088175 \\
SINet & C10 & 42 & NC4K & 0.765296 & 0.086273 \\
SINet & C10 & 43 & NC4K & 0.771241 & 0.085305 \\
SINet & C10 & 45 & NC4K & 0.770503 & 0.086622 \\
SINet & CINV & 42 & NC4K & 0.774697 & 0.083144 \\
SINet & CINV & 43 & NC4K & 0.774114 & 0.084030 \\
SINet & CINV & 45 & NC4K & 0.768136 & 0.084570 \\
SINet & CSHUF & 42 & NC4K & 0.766713 & 0.085734 \\
SINet & CSHUF & 43 & NC4K & 0.767983 & 0.085825 \\
SINet & CSHUF & 45 & NC4K & 0.767229 & 0.089734 \\
SINet & MT & 42 & NC4K & 0.752869 & 0.108067 \\
SINet & MT & 43 & NC4K & 0.746708 & 0.101192 \\
SINet & MT & 45 & NC4K & 0.749400 & 0.104323 \\
SINetv2 & A0 & 42 & NC4K & 0.745691 & 0.092348 \\
SINetv2 & A0 & 43 & NC4K & 0.736067 & 0.099838 \\
SINetv2 & A0 & 45 & NC4K & 0.751473 & 0.092634 \\
SINetv2 & A2 & 42 & NC4K & 0.747102 & 0.093372 \\
SINetv2 & A2 & 43 & NC4K & 0.743257 & 0.097429 \\
SINetv2 & A2 & 45 & NC4K & 0.753292 & 0.090255 \\
SINetv2 & B & 42 & NC4K & 0.753317 & 0.097475 \\
SINetv2 & B & 43 & NC4K & 0.743653 & 0.095901 \\
SINetv2 & B & 45 & NC4K & 0.748868 & 0.091476 \\
SINetv2 & C10 & 42 & NC4K & 0.748534 & 0.096669 \\
SINetv2 & C10 & 43 & NC4K & 0.750839 & 0.097324 \\
SINetv2 & C10 & 45 & NC4K & 0.749305 & 0.092402 \\
SINetv2 & CINV & 42 & NC4K & 0.752956 & 0.095453 \\
SINetv2 & CINV & 43 & NC4K & 0.753600 & 0.089466 \\
SINetv2 & CINV & 45 & NC4K & 0.753833 & 0.094549 \\
SINetv2 & CSHUF & 42 & NC4K & 0.754154 & 0.093821 \\
SINetv2 & CSHUF & 43 & NC4K & 0.750543 & 0.094270 \\
SINetv2 & CSHUF & 45 & NC4K & 0.749500 & 0.095273 \\
\bottomrule
\end{tabular}
\end{center}
\end{table}

\begin{table}[h]
\caption{The ten most-funded clusters under Equation~\ref{eq:alloc} ($N = 1000$, $\atemp = 1.0$), and what the budget is actually aimed at. The target pool is $24.8\%$ CAMO, but $51.5\%$ of the generation budget lands on CAMO images --- a $2.08\times$ enrichment ($2.14\times$ under the committed assignment). The two largest quotas take $33.5\%$ of the budget, are $90.2\%$ CAMO, and between them hold $7$ of the $2026$ primary-endpoint images. \emph{Source:} \texttt{rebuild/DIAG/out/diag\_clusters.json}.}
\label{tab:clustercomp}
\begin{center}
\small
\begin{tabular}{rrrrrrr}
\toprule
Cluster & Budget & \% of $N$ & $n_{\text{target}}$ & CAMO \% & $n_{\text{test}}$ & $e_j$ \\
\midrule
30 & 194 & 19.4 & 21 & 95.2 & 2 & 0.0707 \\
34 & 141 & 14.1 & 81 & 85.2 & 5 & 0.0674 \\
63 & 74 & 7.4 & 18 & 66.7 & 3 & 0.0609 \\
13 & 67 & 6.7 & 57 & 17.5 & 24 & 0.0599 \\
23 & 29 & 2.9 & 43 & 41.9 & 20 & 0.0515 \\
73 & 26 & 2.6 & 39 & 46.2 & 16 & 0.0505 \\
65 & 24 & 2.4 & 64 & 51.6 & 23 & 0.0496 \\
19 & 23 & 2.3 & 93 & 8.6 & 66 & 0.0490 \\
0 & 22 & 2.2 & 57 & 1.8 & 31 & 0.0485 \\
64 & 19 & 1.9 & 141 & 29.1 & 74 & 0.0471 \\
\midrule
\multicolumn{7}{l}{\emph{Budget-weighted totals against the pool they are drawn from}} \\
\multicolumn{4}{l}{CAMO share of the target pool} & \multicolumn{3}{r}{$0.248$} \\
\multicolumn{4}{l}{CAMO share of the \emph{budget}} & \multicolumn{3}{r}{$\mathbf{0.515}$} \\
\multicolumn{4}{l}{Endpoint images per unit budget, against uniform} & \multicolumn{3}{r}{$0.722$} \\
\bottomrule
\end{tabular}
\end{center}
\end{table}

\clearpage
\section{Coverage of the Target Manifold, and Why AUC Is Not Used}
\label{app:coverage}

Section~\ref{sec:why} measures how much of the real target manifold a set covers as $k$-NN recall
at $k = 5$~\cite{kynkaanniemi2019improved}: the fraction of the $4040$ target images that fall inside
the $k$-NN balls of the compared set, in each of the three embedding spaces of Table~\ref{tab:silhouette}.
Every real set covers the target near its own ceiling; both synthetic pools collapse, most sharply in
\texttt{clipL224}.

\begin{center}\small
\begin{tabular}{lccc}
\toprule
Compared set & \texttt{dinoL224} & \texttt{dinoL518} & \texttt{clipL224} \\
\midrule
target against itself (random halves) & 0.9332 & 0.9431 & 0.8931 \\
COD10K part vs CAMO part of the target & 0.8898 & 0.9003 & 0.8730 \\
NC4K, a different real set & 0.8874 & 0.8995 & 0.7921 \\
raw HKU-IS, the generator's input & 0.7473 & 0.7097 & 0.7470 \\
authors' synthetic pool & 0.4829 & 0.4834 & 0.1277 \\
our synthetic renders & 0.4668 & 0.5421 & 0.1349 \\
\bottomrule
\end{tabular}
\end{center}

A real-versus-synthetic classifier is not used as evidence, because it cannot distinguish a
distributional gap from a re-encoding artifact: re-saving the \emph{identical} target images at
JPEG quality $30$ already separates them from themselves at AUC $0.9928$ in \texttt{clipL224}, against
$0.9993$ and $0.9995$ for the two synthetic pools. \emph{Source:} \texttt{rebuild/A3/out/a3\_coverage.csv},
\texttt{a3\_probe\_table.csv}.

\section{CHAMELEON Contamination: Calibration, Controls and Adjudication}
\label{app:chamext}

\paragraph{Tier A--the released same-dimension detector.} Images are grouped by exact pixel
dimensions, since a re-encode preserves them. Within a group, each image is reduced to a
$32\times32$ greyscale descriptor, pairs are shortlisted by descriptor RMS distance at $14.0$, and
survivors are decoded at full resolution and confirmed when the mean absolute difference over RGB
falls below $6/255$. We deliberately do \emph{not} contrast-normalize the descriptor: normalizing
destroys the scale that separates a re-encode from a merely similar photograph, and it is what made
our own first pass report $10$ of $76$ rather than $41$. Sweeping the tolerance over
$1, 2, 3, 5, 6$ gives $11$, $26$, $37$, $40$ and $41$, the count saturates. The detector passes an
eight-assertion synthetic known-answer test that uses no repository data.

\paragraph{Tier B-- calibration of the geometric extension.} SIFT correspondences, Lowe ratio
$0.75$, filtered by a RANSAC homography at $4$ px. Thresholds are fixed on data whose answer is
already committed, before the unchecked images are scored: positives are the $41$ Tier-A pairs,
negatives $4000$ random CHAMELEON$\times$training pairs.

\begin{center}\small
\begin{tabular}{lc}
\toprule
Inliers on the $41$ confirmed pairs & $525$--$1174$ (median $1129$) \\
Inliers on $4000$ random pairs & max $8$ \\
Inliers anywhere in the NC4K control & max $19$ \\
\textbf{Operating point} & \textbf{$20$} \\
Recall on the knowns at that point & $41/41$ \\
NC4K flagged at that point & $0$ of $120$ \\
\bottomrule
\end{tabular}
\end{center}

The operating point is one inlier above the largest count \emph{any} negative set produced, which is
the NC4K maximum of $19$ rather than the random-pair maximum of $8$. We also ran resolution-normalized
global matching as a second instrument; no unchecked image passed it (lowest normalized RMS $16.7$
against a calibrated threshold of $1.02$), so every Tier-B match rests on the keypoint stage alone.

\paragraph{Adjudication from ``the keypoints agree'' to ``the same photograph''.} An inlier count
shows that local patches agree on a geometric model. Two exposures of one static scene agree on a
homography as well, so the inlier count alone cannot carry the claim that these files are the same
photograph. For each candidate, we recover the homography, warp the partner into the CHAMELEON frame, measure the residual over the valid overlap, and we decompose the homography into scale,
rotation, shear, and perspective. Run on $8$ of the $41$ known pairs, the instrument returns their
recorded mean absolute difference to within $0.002$ ($0.603 \to 0.603$, $5.512 \to 5.514$) at
$H = I$, which is what licenses reading its output on the rest.

\begin{center}\small
\setlength{\tabcolsep}{4pt}
\begin{tabular}{lrrrrl}
\toprule
CHAMELEON & Inliers & Residual & Corr. & Scale & Partner pool \\
\midrule
\texttt{animal-16} & $2015$ & $3.76$ & $0.99$ & $1.02$ & CAMO \\
\texttt{animal-73} & $3332$ & $3.93$ & $1.00$ & $1.00$ (crop) & COD10K-train \\
\texttt{animal-74} & $3403$ & $4.17$ & $1.00$ & $1.00$ (crop) & COD10K-train \\
\texttt{animal-22} & $3097$ & $5.84$ & $1.00$ & $1.00$ (crop) & COD10K-train \\
\texttt{animal-72} & $396$ & $6.60$ & $0.99$ & $6.36$ & CAMO \\
\texttt{animal-61} & $1456$ & $8.92$ & $0.97$ & $0.98$ & CAMO \\
\texttt{animal-43} & $263$ & $9.54$ & $0.96$ & $0.53$ & CAMO \\
\texttt{animal-66} & $860$ & $10.72$ & $0.97$ & $0.51/0.39$ & COD10K-train \\
\texttt{animal-59} & $1017$ & $12.11$ & $0.97$ & $1.52$ & CAMO \\
\midrule
\texttt{animal-7} & $25$ & \multicolumn{4}{l}{\textbf{rejected}; shear $-86.5^\circ$, scale $0.24$ vs $1.24$; no homography at full resolution} \\
\bottomrule
\end{tabular}
\end{center}

Residual is mean $|{\cdot}|$ over the overlap on a $0$--$255$ scale. Shear and perspective are
indistinguishable from zero for all nine. The $525$--$1174$ band describes same-dimension Tier-A
positives and is not an acceptance rule: rescaled or cropped copies naturally yield fewer inliers
(\texttt{animal-72} and \texttt{animal-43}, at scales $6.36$ and $0.53$, have $396$ and $263$).
\texttt{animal-7} was the only candidate near the operating point, clearing it by $5$ while the NC4K
negatives reached $19$; its homography collapses the frame towards a line, and it is excluded.

\paragraph{Scope of the exact-hash zero.} The $0$ we report is measured against the training pool.
Across benchmarks it is not zero: \texttt{animal-17}, \texttt{animal-25} and \texttt{animal-60} are
byte-identical to CAMO-250 images. Those files are absent from the training pool and so do not enter
the contamination count, but CAMO-250 is the split that this repository selects checkpoints on.

\paragraph{What remains uncounted.} Sixteen images have no partner at either tier and are reported
\textsc{unchecked}, not clean. Different photographs of one specimen, montages, and heavy color edits
are not detected and are not claimed. Every figure here is a lower bound.

\section{CHAMELEON Mask Releases and Detector Validation}
\label{app:masks}

The two circulating CHAMELEON ground-truth releases are different annotations, unrelated to
contamination or to any method. They disagree in polarity, giving a mean IoU of $0.0$ as stored and
$0.6932$ after polarity alignment; only $27$ of $76$ masks are exactly identical after alignment and
$40$ of $76$ reach IoU $\geq 0.9$. On identical predictions, the choice of release alone moves the MAE by
$2.7\times$ ($0.2196$ against the author-sourced masks, $0.0816$ against the repackaged ground truth).
Any CHAMELEON column is therefore not comparable across papers unless the release is named.

\paragraph{Detector self-test.} The released detector passes an eight-assertion synthetic
known-answer test using no repository data, reproduces the canonical $41$ pairs and $41$ names exactly, and surfaced one additional pair, the second partner of the one image that has two.

\section{Three Controls: Execution Records and Disclosures}
\label{app:controls}

Section~\ref{sec:null} closes three objections with three campaigns. Each carries a disclosure that
belongs with it.

\paragraph{The oracle campaign uses test labels, deliberately.} CORACLE takes the committed
allocation's shape and re-assigns it across clusters by the rank of each cluster's \emph{true} error
on the endpoint. That is information no real method could have, and the pre-registration says so
before the first run: it is a diagnostic upper bound and never a method, and a pre-flight gate that
would forbid this for any real arm is passed on a technicality, which we disclose rather than claim
as cleanliness. One further limitation: the oracle is derived from one architecture's errors and
applied to both. The two architectures' per-cluster errors agree at a rank correlation of $0.9211$,
so the second architecture receives a near-oracle rather than a perfect one.

\paragraph{The schedule campaign bundles four changes, not one.} Unpinning the optimization budget
does not isolate a single variable. It raises steps per epoch from $253$ to $341$ (SINet) and $127$
to $171$ (SINet-v2); it re-uses the same $4040$ target images more often, so each is seen about
$1.35$ times per epoch instead of once; it leaves the learning-rate schedule unrescaled, because
rescaling would re-impose a fixed budget and defeat the control; and it therefore changes total
optimization as well as its distribution. Growing the budget by roughly a third made every arm
slightly \emph{worse}, $11$ of $12$ cross-schedule comparisons are negative, most plausibly
because the learning rate decays per epoch and was deliberately left alone. A gate declared in
advance that any run whose step count stayed at the old pinned value was a halt rather than a
result.

\paragraph{The seed expansion is a separate campaign, and survived a power loss.} It is not the
committed campaign with more seeds: adding seeds after seeing a result is optional stopping, and it
would invalidate every verdict the original campaign reports. SE has its own four arms, its own
noise estimate, its own bar, and its own verdicts, and its eight seeds were fixed in writing before
its first run. The twenty-four runs at the committed seeds (twelve per architecture) are re-scored rather than re-trained, and must
reproduce their committed values to six decimal places or the campaign halts.

A host power loss on 2026-09-20 interrupted training. Two runs were killed partway through their
first round; the driver's own pre-launch check discarded both and re-ran them from scratch, and they
are the only two entries in the discard record. No partially trained model entered the scoring pool,
no seed was added or dropped, and no SE number existed at the moment resumption was decided; the
metrics file was absent, so nothing had been computed and nothing seen. The campaign cost $75.9$
GPU-hours across its $40$ trained runs.

\paragraph{What differs between arms beyond the injected images.} The pseudo-labeling step between
rounds selects target images using each arm's \emph{own} first-round model, so each arm appends a
different number of differently-labeled images. Across the seed expansion, the means are $1977$ (B),
$2025$ (C10), $1990$ (CSHUF) and $1979$ (CINV), with per-run values spanning $1817$ to $2259$. This
is legitimately part of ``closed loop'' and is reported rather than absorbed: it is a second place where the arms differ beyond the injection, and it is high-variance. The step count stays pinned
regardless, which is asserted per run.

\end{document}